\documentclass[10pt,twocolumn,letterpaper]{article}

\usepackage{iccv}
\usepackage{times}
\usepackage{epsfig}
\usepackage{graphicx}
\usepackage{amsmath}
\usepackage{amssymb}
\usepackage{xstring}
\usepackage{subfigure}
\usepackage{caption}
\usepackage{float}

\usepackage{pifont}
\newcommand{\cmark}{\ding{51}}%
\newcommand{\xmark}{\ding{55}}%

\usepackage[accsupp]{axessibility}  

\usepackage[pagebackref=true,breaklinks=true,letterpaper=true,colorlinks,bookmarks=false]{hyperref}

\usepackage[capitalize]{cleveref}
\crefname{section}{Sec.}{Secs.}
\Crefname{section}{Section}{Sections}
\Crefname{table}{Table}{Tables}
\crefname{table}{Tab.}{Tabs.}

\iccvfinalcopy 

\def\iccvPaperID{1192} 

\ificcvfinal\fi

\newcommand{\Skip } [1] {}
\newcommand{\zjs}[1]{{\textcolor{magenta}{[zjs:#1]}}}

\usepackage{mathtools}
\newcommand{\defeq}{\vcentcolon=}
\newcommand{\ftext}[1][TODO]{\textcolor[rgb]{0.5, 0.56, 1}{#1}}
\newcommand{\fcite}[1]{[\ftext[#1]]}

\newcommand{\msymbol}[1]{%
    \IfEqCase{#1}{%
        {point_pos}{\mathbf{x}}
        {point_dir}{\mathbf{d}}
        {point_color}{c}
        {point_sigma}{\sigma}
        {pixel_color}{C}
        {pixel_depth}{D}
        {point_color_mapped}{c^{T}}
        {point_sigma_mapped}{\sigma^{T}}
        {pixel_color_edited}{\hat{C}^{T}}
        {pixel_depth_edited}{\hat{D}^{T}}
        {point_color_student}{c^{S}}
        {point_sigma_student}{\sigma^{S}}
        {pixel_color_student}{\hat{C}^{S}}
        {pixel_depth_student}{\hat{D}^{S}}
        {infer_func}{F}
        {mapper_func}{F^{m}}%
        {points_raw}{B}%
        {points_mapped}{P^{t}}%
        {teacher_model}{M^{t}}%
        {student_model}{M^{s}}%
        {loss_pretrain}{\mathcal{L}_\mathrm{local}}
        {weight_pretrain_color}{\lambda_1}
        {weight_pretrain_sigma}{\lambda_2}
        {loss_train}{\mathcal{L}_\mathrm{global}}
        {weight_train_color}{\lambda_3}
        {weight_train_depth}{\lambda_4}
    }[\PackageError{msymbol}{Undefined option to msymbol: #1}{}]%
}%

\renewcommand{\eqref}[2][\reflabel]{(\ref{eq:#1-#2})}

\newcommand{\be}{\begin{equation}}
\newcommand{\ee}{\end{equation}}

\begin{document}

\newcommand{\yq}[1]{\textcolor{red}{\textbf{yq: }\xspace#1}\xspace}
\title{Seal-3D: Interactive Pixel-Level Editing for Neural Radiance Fields}

\author{Xiangyu Wang\textsuperscript{1}\thanks{Equal contribution.}
\qquad
Jingsen Zhu\textsuperscript{2}\footnotemark[1]
\qquad
Qi Ye\textsuperscript{1}\thanks{Corresponding author.}
\qquad
Yuchi Huo\textsuperscript{3,2}
\qquad
Yunlong Ran\textsuperscript{1}
\qquad
\\
Zhihua Zhong\textsuperscript{2}
\qquad
Jiming Chen\textsuperscript{1}
\qquad
\smallskip
\and
\textsuperscript{1}Zhejiang University, Key Lab of CS\&AUS of Zhejiang Province
\\
\textsuperscript{2}State Key Lab of CAD\&CG, Zhejiang University
\qquad
\textsuperscript{3}Zhejiang Lab
\smallskip
\\
\tt{\small\{\href{mailto:xy_wong@zju.edu.cn}{\textcolor{black}{xy\_wong}}, \href{mailto:zhujingsen@zju.edu.cn}{\textcolor{black}{zhujingsen}}, \href{mailto:qi.ye@zju.edu.cn}{\textcolor{black}{qi.ye}}\}@zju.edu.cn}
\qquad
\tt{\small \href{mailto:huo.yuchi.sc@gmail.com}{\textcolor{black}{huo.yuchi.sc@gmail.com}}}
\\
\tt{\small\{\href{mailto:yunlong_ran@zju.edu.cn}{\textcolor{black}{yunlong\_ran}}, \href{mailto:zhongzhihua@zju.edu.cn}{\textcolor{black}{zhongzhihua}}, \href{mailto:cjm@zju.edu.cn}{\textcolor{black}{cjm}}\}@zju.edu.cn}
}

\ificcvfinal\thispagestyle{empty}\fi

\twocolumn[{%
\renewcommand\twocolumn[1][]{#1}%
\maketitle
\begin{center}
    \centering
    \captionsetup{type=figure}
    \vspace{-3em}
    \includegraphics[width=\textwidth]{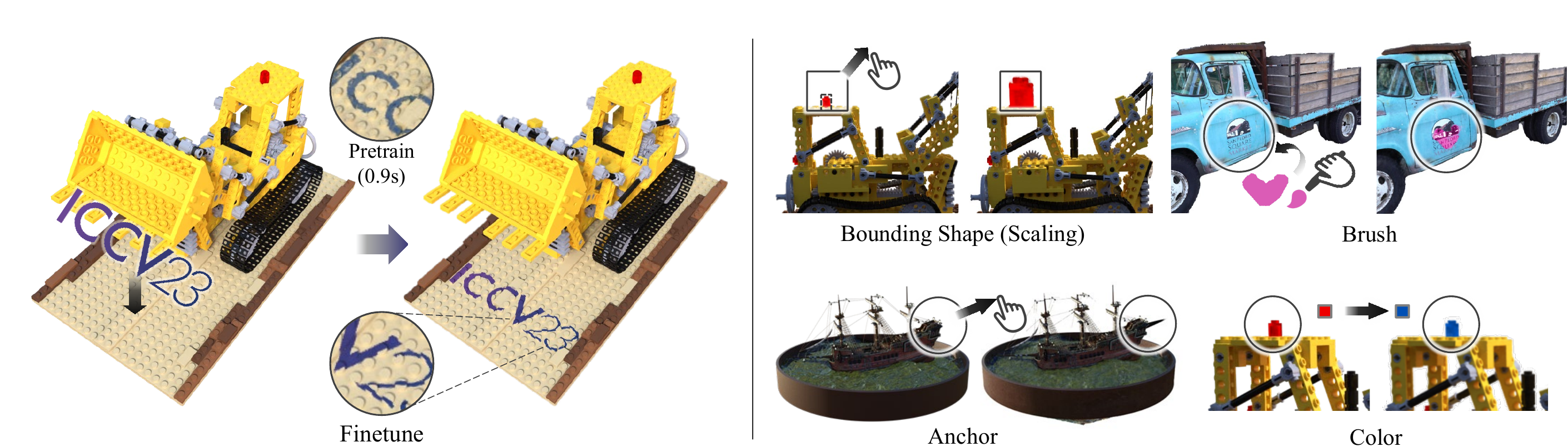}
    \captionof{figure}{Seal-3D: The first interactive pixel level NeRF editing tool. We design an interactive user editing method and system \textit{Seal-3D}, which achieves instant ($\approx$1s) preview (left) by our novel pretraining strategy. High-quality editing results can be further obtained by a short period (in 1 or 2 minutes) of finetuning. The editing results of our implemented editing tools (right) are view-consistent with rich shading details (\eg shadows) on the original surface (left).}
    \label{fig-teaser}
\end{center}%
}]

\newcommand\blfootnote[1]{%
  \begingroup
  \renewcommand\thefootnote{}\footnote{#1}%
  \addtocounter{footnote}{-1}%
  \endgroup
}
\blfootnote{
$^*$Equal contribution.
}
\blfootnote{
$^\dagger$Corresponding author. }
\blfootnote{
Project page: \href{https://windingwind.github.io/seal-3d/}{https://windingwind.github.io/seal-3d/}
}

\begin{abstract}
With the popularity of implicit neural representations, or neural radiance fields (NeRF), there is a pressing need for editing methods to interact with the implicit 3D models for tasks like  post-processing reconstructed scenes and 3D content creation. While previous works have explored NeRF editing from various perspectives, they are restricted in editing flexibility, quality, and speed, failing to offer direct editing response and instant preview. The key challenge is to conceive a locally editable neural representation that can directly reflect the editing instructions and update instantly.

To bridge the gap, we propose a new interactive editing method and system for implicit representations, called Seal-3D
\footnote{``Seal'' derived from the name of rubber stamp in Adobe Photoshop.}
, which allows users to edit NeRF models in a pixel-level and free manner with a wide range of NeRF-like backbone and preview the editing effects instantly. To achieve the effects, the challenges are addressed by our proposed proxy function mapping the editing instructions to the original space of NeRF models in the teacher model and a two-stage training strategy for the student model with local pretraining and global finetuning. A NeRF editing system is built to showcase various editing types. Our system can achieve compelling editing effects with an interactive speed of about 1 second.
\end{abstract}


\section{Introduction}


Implicit neural representations, \eg neural radiance fields (NeRF)~\cite{mildenhall2020nerf}, have gained increasing attention as novel 3D representations with neural networks to model a 3D scene. Benefiting from the high reconstruction accuracy and rendering quality with relatively low memory consumption, NeRF and its variations~\cite{kaizhang2020,barron2021mipnerf,yu_and_fridovichkeil2021plenoxels,mueller2022instant,Chen2022ECCV,yariv2021volume,wang2021neus}
have demonstrated great potential in many 3D applications like 3D reconstruction, novel view synthesis, and Virtual/Augmented Reality.

With the popularity of the new implicit representations and an increasing number of implicit 3D models, there is a pressing demand for human-friendly editing tools to interact with these 3D models. Editing with implicit neural representations is a fundamental technique required to fully empower the representation. Objects reconstructed from the real world are likely to contain artifacts due to the noise of captured data and the limitations of the reconstruction algorithms. In a typical 3D scanning pipeline, manual correction and refinement to remove artifacts are common stages. On the other hand, in 3D content creation applications like 3D games, animations, and filming, artists usually need to create new content based on existing 3D models. 

Prior works have made attempts to edit 3D scenes represented by NeRF, including object segmentation~\cite{liu2021editing,yang2021objectnerf}
, object removal~\cite{liu2022nerf}
, appearance editing~\cite{kuang2022palettenerf,munkberg2022extracting,mikaeili2023sked}
, and object blending~\cite{guo2021template}, \etc. These existing NeRF editing methods mainly focus on coarse-grained object-level editing and the convergence speed can not meet the demands of interactive editing. Some recent methods~\cite{Yuan22NeRFEditing,neumesh} transform the editing of NeRF into mesh editing by introducing a mesh as an edit proxy. This requires the user to operate on an additional meshing tool, which limits interactivity and user-friendliness. To the best of our knowledge, there are no existing methods that are able to support interactive pixel-level editing of neural radiance fields with fast converging speed, which is mainly due to the challenges discussed below.

Unlike existing explicit 3D representations \eg point cloud, textured mesh, and occupancy volume, which store the explicit geometry structure of objects and scenes, implicit representations use neural networks to query features of a 3D scene including geometry and color. Existing 3D editing methods, taking the mesh-based representation as an example, can change object geometry by displacing vertices corresponding to target object surface areas and object textures. Without explicit explainable correspondence between the visual effects and the underlying representations, editing the implicit 3D models is indirect and challenging. 
Further, it is difficult to locate implicit network parameters in local areas of the scene,
meaning that adaptations of the network parameters may lead to undesired global changes. This results in more challenges for fine-grained editing.

To bridge the gap, in this paper, we propose an interactive pixel-level editing method and system for implicit neural representations for 3D scenes, dubbed Seal-3D. The name is borrowed from the popular 2D image editing software Adobe PhotoShop~\cite{adobephotoshop}, as its seal tool provides similar editing operations. As shown in \cref{fig-teaser}, the editing system consists of five types of editing as examples: 1) Bounding box tool. It transforms and scales things inside a bounding box, like a copy-paste operation. 2) Brushing tool. It paints specified color on the selected zone and can increase or decrease the surface height, like an oil paint brush or graver. 3) Anchor tool. It allows the user to freely move a control point and affect its neighbor space according to the user input. 4) Color tool. It edits the color of the object surfaces.

To achieve the interactive NeRF editing effects, we address the challenges of implicit representations discussed above. First, to establish the correspondence between the explicit editing instructions to the update of implicit network parameters, we propose a proxy function that maps the target 3D space (determined by the user edit instructions from an interactive GUI) to the original 3D scene space, and a teacher-student distillation strategy to update the parameters with the corresponding content supervision acquired by the proxy function from the original scenes. Second, to enable local editing, \ie mitigating the influence of the local editing effect on the global 3D scenes under the non-local implicit representations, we propose a two-stage training process: 
a pretraining stage of updating only the positional embedding grids with local losses for editing areas while freezing the subsequent MLP decoder to prevent global degeneration, and a finetuning stage of updating both the embedding grids and the MLP decoder with global photometric losses. With this design, the pretraining stage updates local editing features and the finetuning stage blends the local editing areas with global structures and colors of unedited space to achieve view consistency. This design has the benefit of an instant preview of the editing: the pretraining can converge very fast and presents local editing effects within approximately 1 second only.

In summary, our contributions are as follows:

\begin{itemize}
\item We propose the first interactive pixel-level editing method and system for neural radiance fields, which exemplifies fine-grained multiple types of editing tools, including geometry (bounding box tool, brush tool, and anchor tool) and color edits;
\item A proxy function is proposed to establish the correspondence between the explicit editing instructions and the update of implicit network parameters and a teacher-student distillation strategy is proposed to update the parameters;
\item A two-stage training strategy is proposed to enable instant preview of local fine-grained editing without contaminating the global 3D scenes. 
\end{itemize}

\section{Related Work}

\begin{figure*}[ht]
\centering
    \includegraphics[width=\linewidth]{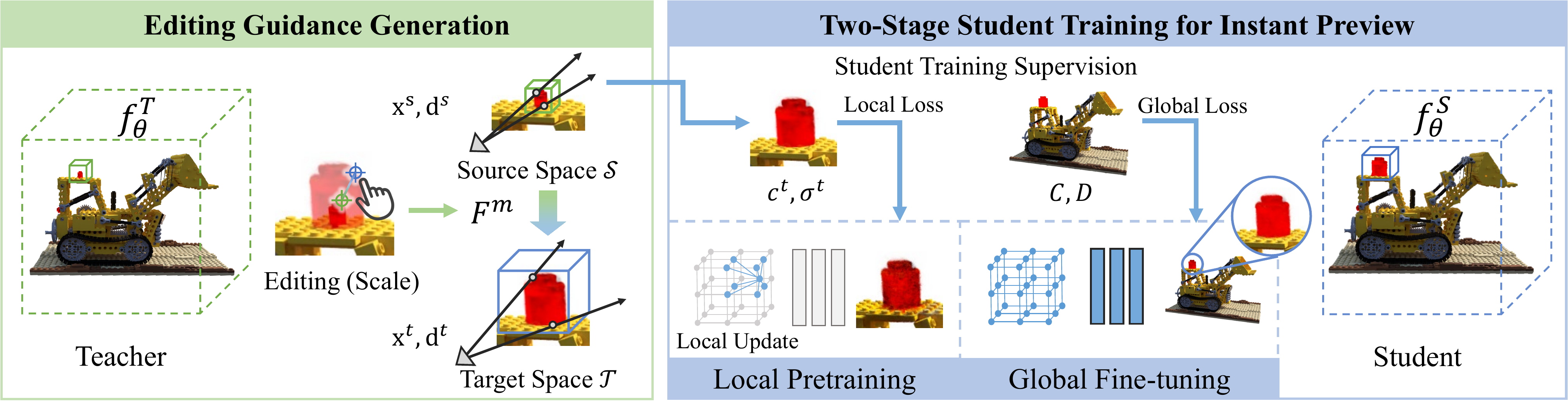}
    \caption{Illustration of the editing framework. Left: a 3D point and view direction from the target space after user editing is mapped to the original source space to get guidance ${c}_t, \sigma_t$ from the teacher model $f_\theta^T$  for the student training. Right: the student training consists of two stages: fast pretraining to provide instant preview by updating partial parameters of the network with local losses and finetuning with global losses.   }    
    \label{fig-framework}
\end{figure*}

\paragraph{Novel view synthesis.} Given a set of posed image captures of a scene, the task of novel view synthesis is to generate photo-realistic images from arbitrary novel views. Recently, neural network have been introduced into the rendering pipeline and leveraged for multiple representations, such as voxels~\cite{Lombardi2019,sitzmann2019deepvoxels}, point clouds~\cite{Aliev2020,dai2020neural}, multi-plane images (MPIs)~\cite{li2020crowdsampling,mildenhall2019llff,zhou2018stereo}, and implicit representations~\cite{sitzmann2019srns,mildenhall2020nerf}. Typically, Neural radiance field (NeRF)~\cite{mildenhall2020nerf} uses a single MLP to implicitly encode a scene into a volumetric field of density and color, and takes advantage of volume rendering to achieve impressive rendering results with view-dependent effects, which inspires a lot of follow-up works on human~\cite{peng2021neural,weng_humannerf_2022_cvpr}, deformable objects~\cite{park2021nerfies,park2021hypernerf}, pose estimations~\cite{lin2021barf}, autonomous system~\cite{ran2023neurar,zeng2023effic}, surface reconstruction~\cite{yariv2021volume,wang2021neus}, indoor scenes~\cite{Yu2022MonoSDF}, city~\cite{tancik2022blocknerf,xiangli2022bungeenerf}, \etc. NeRF's MLP representation can be enhanced and accelerated by hybrid representations, including voxels~\cite{yu_and_fridovichkeil2021plenoxels,SunSC22}, hashgrids~\cite{mueller2022instant} and tensorial decomposition~\cite{Chen2022ECCV,tang2022compressible}. In this paper, our interactive editing framework is developed based on Instant-NGP~\cite{mueller2022instant}, which achieve real-time rendering speed for NeRF inference and state-of-the-art quality of novel view synthesis.

\paragraph{Neural scene editing.} Scene editing has been a widely researched problem in computer vision and graphics. Early method focus on editing a single static view by inserting~\cite{li2020inverse,zhu2022learning}, relighting~\cite{li2022physically}, composition~\cite{perez2003poisson}, object moving~\cite{OM3D2014,shetty_neurips2018}, \etc. With the development of neural rendering, recent works attempt to perform editing at different levels of the 3D scene, which can be categorized as scene-level, object-level, and pixel-level editing. Scene-level editing methods focus on changing in global appearances of a scene, such as lighting~\cite{guo2020object} and global palette~\cite{kuang2022palettenerf}. Intrinsic decomposition~\cite{zhang2021nerfactor,munkberg2022extracting,hasselgren2022nvdiffrecmc,Ye2022IntrinsicNeRF,zhu2023i2,10.1145/3588432.3591493} disentangles material and lighting field and enables texture or lighting editing. However, scene-level methods are only able to modify global attributes and are unable to apply to specified objects. Object-level editing methods use different strategies to manipulate the implicitly represented object. Object-NeRF~\cite{yang2021objectnerf} exploit per-object latent code to decompose neural radiance field into objects, enabling object moving, removal, or duplicating. Liu \etal~\cite{liu2021editing} design a conditional radiance field model which is partially optimized according to the editing instructions to modify semantic-level color or geometry. NeRF-editing~\cite{Yuan22NeRFEditing} and NeuMesh~\cite{neumesh} introduce a deformable mesh reconstructed by NeRF, as an editing proxy to guide object editings. However, these methods are restricted to object-level rigid transformation or are not generalizable to arbitrary out-of-distribution editing categories. In contrast, pixel-level editing aims to provide fine-grained editing guidance precisely selected by pixels, instead of restricted by object entities. To the best of our knowledge, NeuMesh~\cite{neumesh} is the only existing method that achieves editing at this level. However, it depends on the mesh scaffold, which limits the editing categories, \eg cannot create out-of-mesh geometry structures. In contrast, our editing framework does not require any proxy geometry structures, allowing it to be more direct and extensive.

Besides, optimizing the performance of neural editing method remains an open problem. Existing methods require minutes or even hours of optimization and inference. Our method is the first pixel-level neural editing framework to achieve instant interactive (\ie second-level) performance.

\section{Method}

We introduce Seal-3D, an interactive pixel-level editing method for neural radiance fields. The overall pipeline is illustrated in \cref{fig-framework}, which consists of a pixel-level proxy mapping function, a teacher-student training framework, and a two-stage training strategy for the student NeRF network under the framework. Our editing workflow starts with the proxy function which maps the query points and ray directions according to user-specified editing rules. Then a NeRF-to-NeRF teacher-student distillation framework follows, where a teacher model with editing mapping rules of geometry and color supervises the training of a student model (\cref{sec-teacher}).  
The key to interactive fine-grained editing is the two-stage training for the student model (\cref{sec-train}). In an extra pretraining stage, the points, ray directions, and inferred ground truth inside edit space from the teacher model are sampled, computed, and cached previously; only parameters with locality are updated and the parameters causing global changes are frozen. After the pretraining stage, the student model is finetuned with a global training stage.

\subsection{Overview of NeRF-based Editing Problem}
We first make a brief introduction to neural radiance fields and then analyze the challenges of NeRF-based editing problems and the limitations of existing solutions.

\subsubsection{NeRF Preliminaries}

Neural radiance fields (NeRFs) provide implicit representations for a 3D scene as a 5D function: $f: (x,y,z,\theta,\varphi)\mapsto(c,\sigma)$, where $\mathbf{x}=(x,y,z)$ is a 3D location and $\mathbf{d}=(\theta,\phi)$ is the view direction, while $c$ and $\sigma$ denote color and volume density, respectively. The 5D function is typically parameterized as an MLP $f_\theta$.

To render an image pixel, a ray $\mathbf{r}$ with direction $\mathbf{d}$ is shot from the camera position $\mathbf{o}$ through the pixel center according to the intrinsics and extrinsics of the camera. $K$ points $\mathbf{x}_i = \mathbf{o} + t_i\mathbf{d}, i=1,2,\ldots,K$ are sampled along the ray, and the network $f_\theta$ is queried for their corresponding color and density:
\begin{equation}
    (c_i,\sigma_i) = f_\theta(\mathbf{x}_i,\mathbf{d})
\end{equation}
Subsequently, the predicted pixel color $\hat{C}(\mathbf{r})$ and depth value $\hat{D}(\mathbf{r})$ are computed by volume rendering:
\begin{align}
    \hat{C}(\mathbf{r}) &= \sum_{i=1}^K{T_i\alpha_i{c}_i}, &\hat{D}(\mathbf{r}) &=  \sum_{i=1}^K{T_i\alpha_i t_i} \label{eq-render}\\
    T_i &= \prod_{j<i}(1-\alpha_j), &\alpha_i &= 1 - \exp{(\sigma_i\delta_i)}
\end{align}
where $\alpha_i$ is the alpha value for blending, $T_i$ is the accumulated transmittance, and $\delta_i = t_{i+1} - t_i$ is the distance between adjacent points. NeRF is trained by minimizing the photometric loss between the predicted and ground truth color of pixels.



In this paper, we build our interactive NeRF editing system upon Instant-NGP~\cite{mueller2022instant}, which achieves nearly real-time rendering performance for NeRF. Although our implementation of instant interactive editing relies on hybrid representations for NeRF
to achieve the best speed performance, our proposed editing framework does not rely on a specific NeRF backbone and can be transplanted to other frameworks as long as they follow the aforementioned volume rendering pipeline.

\subsubsection{Challenges of NeRF-based Editing}

NeRF-like methods achieve the state-of-the-art quality of scene reconstruction. However, the 3D scene is implicitly represented by network parameters, which lacks interpretability and can hardly be manipulated. In terms of scene editing, it is difficult to find a mapping between the \textit{explicit} editing instructions and the \textit{implicit} update of network parameters. Previous works attempt to tackle this by means of several restricted approaches:


NeRF-Editing~\cite{Yuan22NeRFEditing} and NeuMesh~\cite{neumesh} introduce a mesh scaffold as a geometry proxy to assist the editing, which simplifies the NeRF editing task into mesh modification. Although conforming with existing mesh-based editing, the editing process requires extracting an additional mesh, which is cumbersome.  In addition, the edited geometry is highly dependent on the mesh proxy structure, making it difficult to edit spaces that are not easy or able to be represented by meshes while representing these spaces is one key feature of the implicit representations.
Liu \etal~\cite{liu2021editing} designs additional color and shape losses to supervise the editing. However, their designed losses are only in 2D photometric space, which limits the editing capability of a 3D NeRF model. Furthermore, their method only supports editing of semantic-continuous geometry in simple objects, instead of arbitrary pixel-level complex editing.




Moreover, to the best of our knowledge, existing methods have not realized interactive editing performance considering both quality and speed. Liu \etal~\cite{liu2021editing} is the only existing method that completes optimization within a minute (37.4s according to their paper), but their method only supports extremely simple objects and does not support fine-grained local edits (see \cref{fig-editnerf} for details).
Other editing methods (\eg NeuMesh~\cite{neumesh}) usually require hours of network optimization to obtain edit results. 

In this paper, we implement an interactive pixel-level editing system, which can be extended to new editing types easily using similar editing strategies as the traditional explicit 3D representation editing. Our method does not require any explicit proxy structure (instead, a proxy function, see \cref{sec-teacher}) and can define various pixel-level editing effects without an explicit geometry proxy. It also enables \emph{instant preview} ($\approx$1s) (see \cref{sec-train}). \cref{tab-baseline} compares the edit capabilities between our method and previous methods.


\begin{table}[ht]
    \centering
    \setlength{\tabcolsep}{3pt}
    \small
    \resizebox{\columnwidth}{!}{
    \begin{tabular}{c|cccc}
        Method & w/o Explicit Proxy &  Pixel-Level & Interactive & Time \\\hline
        Ours & \cmark & \cmark & \cmark & seconds \\
        NeuMesh~\cite{neumesh} & \xmark & (partial) & \xmark & hours \\
        NeRF-Editing~\cite{Yuan22NeRFEditing} & \xmark & \xmark & \xmark & hours \\
    \end{tabular}
    }
    \caption{\textbf{Comparison with recent methods in edit capabilities.} Our method supports arbitrary editing, does not require any explicit geometry proxy, and achieves interactive editing in seconds.}
    \label{tab-baseline}
\end{table}

\begin{figure*}[h!]
    \centering
    \vspace{-1em}
    \includegraphics[width=\linewidth, clip]{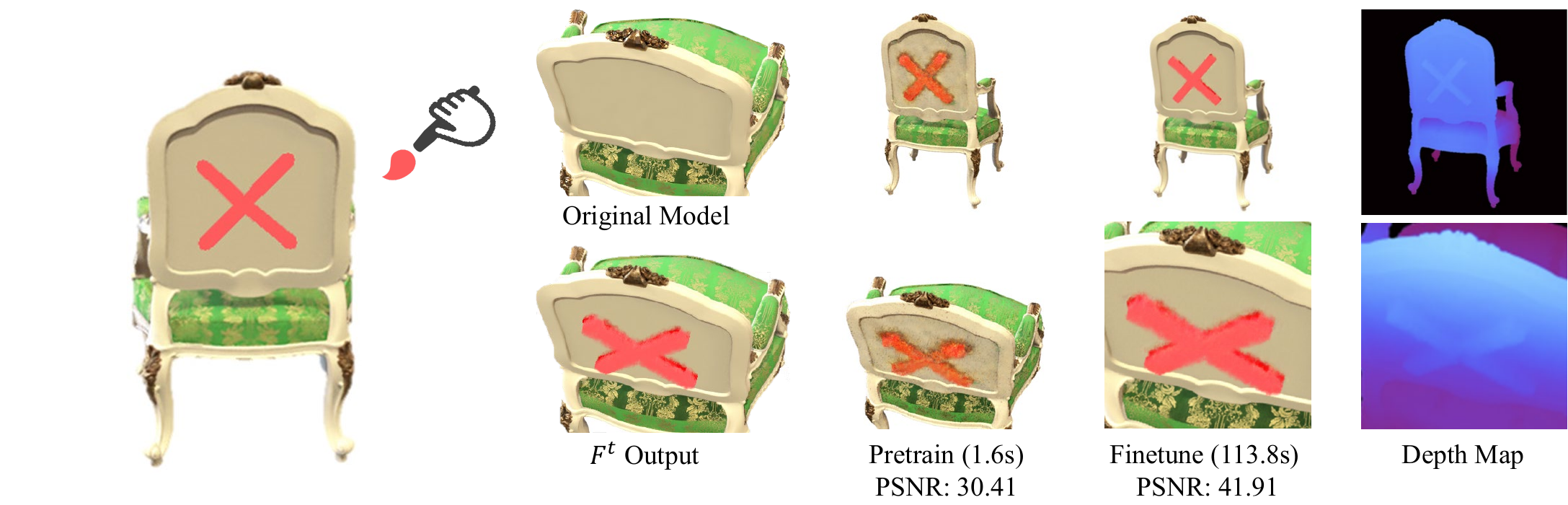}
    \vspace{-1em}
    \caption{Example of brush editing: 3D painting with color and thickness. \Skip{(NeRF Syn. chair) .}}
    \label{fig-brush}
\end{figure*}

\subsection{Editing Guidance Generation}
\label{sec-teacher}

Our design implements NeRF editing as a process of knowledge distillation. Given a pretrained NeRF network fitting a particular scene that serves as a teacher network, we initialize an extra NeRF network with the pretrained weights as a student network. The teacher network $f_\theta^T$ generates editing guidance from the editing instructions input by the user, while the student network $f_\theta^S$ is optimized by distilling editing knowledge from the editing guidance output by the teacher network. In the subsection, editing guidance generation for the student model supervision is introduced and illustrated on the left of \cref{fig-framework}.


Firstly, the user edit instructions are read from the interactive NeRF editor as pixel-level information. The source space $\mathcal{S}\subset\mathbb{R}^3$ is the 3D space for the original NeRF model and the target space $\mathcal{T}\subset\mathbb{R}^3$ is the 3D space for the NeRF model after editing. The target space $\mathcal{T}$ is warped to the original space  $\mathcal{S}$ by  $\msymbol{mapper_func}: \mathcal{T}\mapsto\mathcal{S}$.
$\msymbol{mapper_func}$ transforms points within 
the target space and their associated directions according to editing rules which are exemplified below. 
With the function, the ``pseudo'' desired edited effects  $\msymbol{point_color_mapped}, \msymbol{point_sigma_mapped}$ for each 3D point and view direction in the target space can be acquired by querying the teacher NeRF model $f_\theta^T$: the transformed points and directions (in source space) are fed into the teacher network get the color and density. The process can be expressed as 
\begingroup
\begin{align}
    \mathbf{x}^{s},\mathbf{d}^{s} &= \msymbol{mapper_func}(\mathbf{x}^{t},\mathbf{d}^{t}), \mathbf{x}^{s}\in\mathcal{S}, \mathbf{x}^{t}\in\mathcal{T}, \\
    \msymbol{point_color_mapped},\msymbol{point_sigma_mapped} &= f_\theta^T(\mathbf{x}^{s},\mathbf{d}^{s})
\end{align}
\endgroup

Where $\mathbf{x}^s, \mathbf{d}^s$ denotes source space point position and direction and $\mathbf{x}^t, \mathbf{d}^t$ denotes target space point position and direction.

For brevity, we define the entire process as \textit{teacher inference process} $F^t \defeq f_\theta^T\circ\msymbol{mapper_func}: (\mathbf{x}^{t},\mathbf{d}^{t})\mapsto(\msymbol{point_color_mapped},\msymbol{point_sigma_mapped})$. The inference result $\msymbol{point_color_mapped},\msymbol{point_sigma_mapped}$ mimics the edited scene and acts as the teacher label, the information of which is then distilled by the student network in the network optimization stage.

The mapping rules of $\msymbol{mapper_func}$ can be designed according to arbitrary editing targets. In particular, we implement 4 types of editing as examples.

\begin{itemize}
    \item Bounding shape tool, which supports common features in traditional 3D editing software including copy-paste, rotation, and resizing. The user provides a bounding shape to indicate the original space $\mathcal{S}$ to be edited and rotates, translates, and scales the bounding box to indicate the target effects. The target space $\mathcal{T}$ and mapping function $F^m$ are then parsed by our interface
\begingroup
    \begin{equation}
    \begin{aligned}
        \msymbol{point_pos}^{s} = & S^{-1} \cdot R^T \cdot (\msymbol{point_pos}^{t} - \mathbf{c}^{t})  + \mathbf{c}^{s}, 
        \\ \msymbol{point_dir}^{s} = & R^T \cdot \msymbol{point_dir}^{t}\\
        F^m \defeq & (\msymbol{point_pos}^{t}, \msymbol{point_dir}^{t})
        \mapsto 
        \left\{
        \begin{array}{ll}
            (\msymbol{point_pos}^{s}, \msymbol{point_dir}^{s}) &, \text{if}\ \msymbol{point_pos}^{t}\in \mathcal{T}\\
            (\msymbol{point_pos}^{t}, \msymbol{point_dir}^{t}) &, \text{otherwise}
        \end{array}
        \right.
    \end{aligned}\notag
    \end{equation}
    \endgroup
    where $R$ is rotation, $S$ is scale, and $\mathbf{c}^{s},\mathbf{c}^{t}$ are the center of $\mathcal{S},\mathcal{T}$, respectively. 

    With this tool, we even support cross-scene object transfer, which can be implemented by introducing the NeRF of the transferred object as an additional teacher network in charge of part of the teacher inference process within the target area. We give a result in \cref{fig-bbox-baby}.
    \item Brushing tool, similar to the sculpt brush in traditional 3D editing that lifts or descends the painted surface. The user scribbles with a brush and $\mathcal{S}$ is generated by ray casting on brushed pixels. The brush normal $\mathbf{n}$, and pressure value $\mathrm{p}(\cdot) \in [0, 1]$ are defined by user, which determines the mapping:
    \begin{equation}
    \begin{aligned}
        \msymbol{point_pos}^{s} & =\msymbol{point_pos}^{t} - \mathrm{p}(\msymbol{point_pos}^{t}) \mathbf{n},\\
        F^m &\defeq (\msymbol{point_pos}^{t}, \msymbol{point_dir}^{t}) \mapsto (\msymbol{point_pos}^{s}, \msymbol{point_dir}^{t})
    \end{aligned}\notag
    \end{equation}

\begin{figure*}[ht!]
    \centering
    \includegraphics[width=\linewidth]{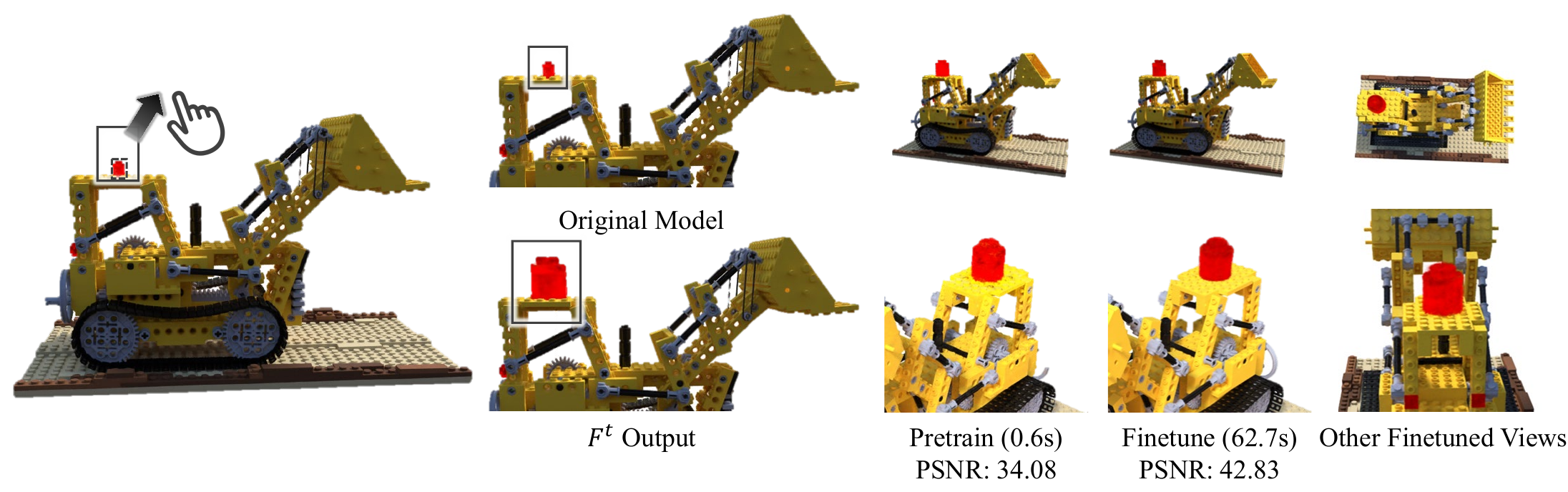}
    \vspace{-1em}
    \caption{Example of bounding shape editing: bulb scaling.}
    \vspace{-1em}
    \label{fig-bbox}
\end{figure*}

    \item Anchor tool, where the user defines a control point $\mathbf{x}^c$ and a translation vector $\mathbf{t}$.
    The region surrounding $\mathbf{x}^c$ will be stretched by a translation function $\mathrm{stretch}(\cdot ;\mathbf{x}^c, \mathbf{t})$. Then the mapping is its inverse:
    \begin{equation}
    \begin{aligned}
        \msymbol{point_pos}^{s} &= \mathrm{stretch}^{-1}(\msymbol{point_pos}^{t};\mathbf{x}^c, \mathbf{t}) \\
        \msymbol{mapper_func} & \defeq (\msymbol{point_pos}^{t}, \msymbol{point_dir}^{t}) \to (\msymbol{point_pos}^{s}, \msymbol{point_dir}^{t})
    \end{aligned}\notag
    \end{equation}
    please refer to the supplementary material for the explicit expressions of $\mathrm{stretch}(\cdot;\mathbf{x}^c, \mathbf{t})$.

    \item Non-rigid transform tool, which allows for the accurate and flexible transformation of selected space.

\begin{equation}
\begin{aligned}
        \mathbf{x}^{s}&=R \cdot \mathbf{x}^{t} + t,\\
        \mathbf{d}^{s}&=R \cdot \mathbf{d}^{t}, \\
    F^m \coloneqq (\mathbf{x}^{t}, \mathbf{d}^{t}) &\mapsto (\mathbf{x}^{s}, \mathbf{d}^{s}) 
\end{aligned}\notag
\end{equation}

\noindent Where $R, t$ are interpolated from the transform matrixes of the three closest coordinates of a pre-defined 3D blending control grid with position and transformation of each control point. The results can be found in \cref{fig-neumesh-mic}.

    \item Color tool, which edits color via color space mapping (single color or texture). Here the spatial mapping is identical and we directly map the color output of the network to HSL space, which helps for color consistency. Our method is capable of preserving shading details (\eg shadows) on the modified surface. We achieve this by transferring the luminance (in HSL space) offsets on the original surface color to the target surface color. Implementation details of this shading preservation strategy are presented in the supplementary.
\end{itemize}



For the training strategy of distillation, the student model $f^S_\theta$ is optimized with the supervision of pseudo ground truths generated by the aforementioned teacher inference process $F^t$. The editing guidance from the teacher model is distilled into the student model by  directly applying the photometric loss between pixel values $\hat{C},\hat{D}$ accumulated by \cref{eq-render} from the teacher and student inference.


However, we find that the convergence speed of this training process is slow ($\approx$30s or longer), which cannot meet the needs of instant preview.  To tackle this problem, we design a two-stage training strategy: the first stage aims to converge instantly (within 1 second) so that a coarse editing result can be immediately presented to the user as a preview, while the second stage further finetunes the coarse preview to obtain a final refinement.

\subsection{Two-stage Student Training for Instant Preview} \label{sec-train}

\paragraph{Local pretraining for instant preview.}
Usually, the edit space is relatively small compared to the entire scene, so training on the global photometric loss is wasteful and leads to slow convergence. To achieve instant preview of editing, we adopt a local pretraining stage before the global training begins. The local pretraining process consists of: 1) uniformly sample a set $\mathcal{X}\subset\mathcal{T}$ of local points within the target space and a set $\mathcal{D}$ of directions on the unit sphere, and feed them into the teacher inference process $F^t$ to obtain teacher labels $\msymbol{point_color_mapped}, \msymbol{point_sigma_mapped}$, and cache them in advance; 2) \Skip{in each pretraining epoch,} the student network is trained by local pertaining loss $\msymbol{loss_pretrain}$:
\begingroup
\begin{align}
    &(\msymbol{point_color_mapped},\msymbol{point_sigma_mapped}) = F^t(\mathbf{x},\mathbf{d}), (\msymbol{point_color_student},\msymbol{point_sigma_student}) = f^S_\theta(\mathbf{x},\mathbf{d}), \\
    &\msymbol{loss_pretrain} = \sum_{\mathbf{x}\in\mathcal{X},\mathbf{d}\in\mathcal{D}}\msymbol{weight_pretrain_color} \|\msymbol{point_color_mapped} - \msymbol{point_color_student}\|_1+ \msymbol{weight_pretrain_sigma} \|\msymbol{point_sigma_mapped} - \msymbol{point_sigma_student}\|_1
\end{align}
\endgroup
where $\msymbol{point_color_student},\msymbol{point_sigma_student}$ are the predicted color and density of sampled points $\mathbf{x}\in\mathcal{X}$ by the student network, and $\msymbol{point_color_mapped}, \msymbol{point_sigma_mapped}$ are cached teacher labels. This pretraining stage is very fast: after only about 1 second of optimization, the rendered image of the student network shows plausible color and shape consistent with the editing instructions.

However, training on only the local points in the editing area may lead to degeneration in other global areas unrelated to the editing due to the non-local implicit neural network. We observe the fact that in hybrid implicit representations (such as Instant-NGP~\cite{mueller2022instant}), local information is mainly stored in the positional embedding grids, while the subsequent MLP decodes global information. Therefore, in this stage, all parameters of the MLP decoder are frozen to prevent global degeneration. Experimental illustrations will be presented in \cref{sec-ablation,fig-ab_fix}.

\paragraph{Global Finetuning.}
After pretraining, we continue to finetune $f^S_\theta$ to refine the coarse preview to a fully converged result. This stage is similar to the standard NeRF training, except that the supervision labels are generated by the teacher inference process instead of image pixels.
\begingroup
\begin{align}
    \msymbol{loss_train} = \sum_{\mathbf{r}\in\mathcal{R}} &\msymbol{weight_train_color} \|\msymbol{pixel_color_edited} - \msymbol{pixel_color_student}\|_2 + \msymbol{weight_train_depth} \|\msymbol{pixel_depth_edited} - \msymbol{pixel_depth_student}\|_1
\end{align}
\endgroup
where $\mathcal{R}$ denote the set of sampled rays in the minibatch and $(\msymbol{pixel_color_edited},\msymbol{pixel_depth_edited})$,$(\msymbol{pixel_color_student},\msymbol{pixel_depth_student})$ are accumulated along ray $\mathbf{r}$ by \cref{eq-render} according to $(\msymbol{point_color_mapped},\msymbol{point_sigma_mapped})$,$(\msymbol{point_color_student},\msymbol{point_sigma_student})$, respectively.

\begin{figure}
\centering
    \includegraphics[width=\linewidth, clip, trim=0 0px 0 0]{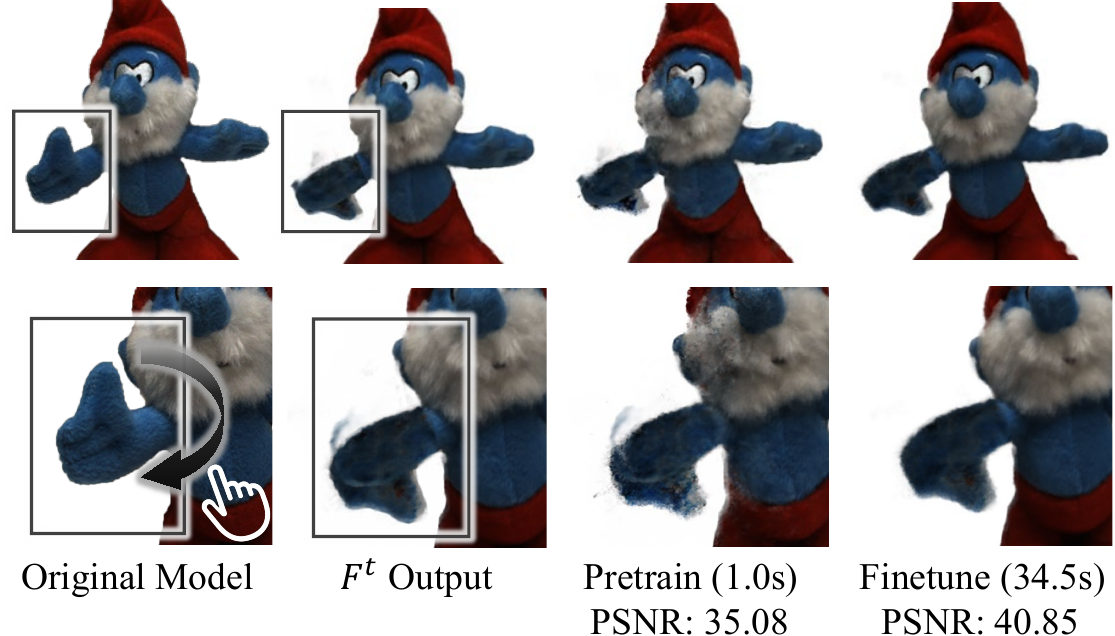}
    \label{fig-realresults-a}
    \vspace{-1.5em}
\caption{Example of editing on the real-world scene: \includegraphics{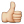} to \includegraphics{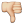}  (DTU Scan 83).}
\label{fig-bbox-elf}
\vspace{-1em}
\end{figure}

It is worth mentioning that the student network is capable of generating results of better quality than the teacher network that it learns from. This is because the mapping operation in the teacher inference process may produce some view-inconsistent artifacts in the pseudo ground truths. However, during the distillation, the student network can automatically eliminate these artifacts due to the multi-view training that enforces view-consistent robustness. See \cref{sec-results,fig-bbox-elf} for details.

\section{Experiments and Analysis}

\begin{figure}
    \centering
    \includegraphics[width=0.9\linewidth, clip]{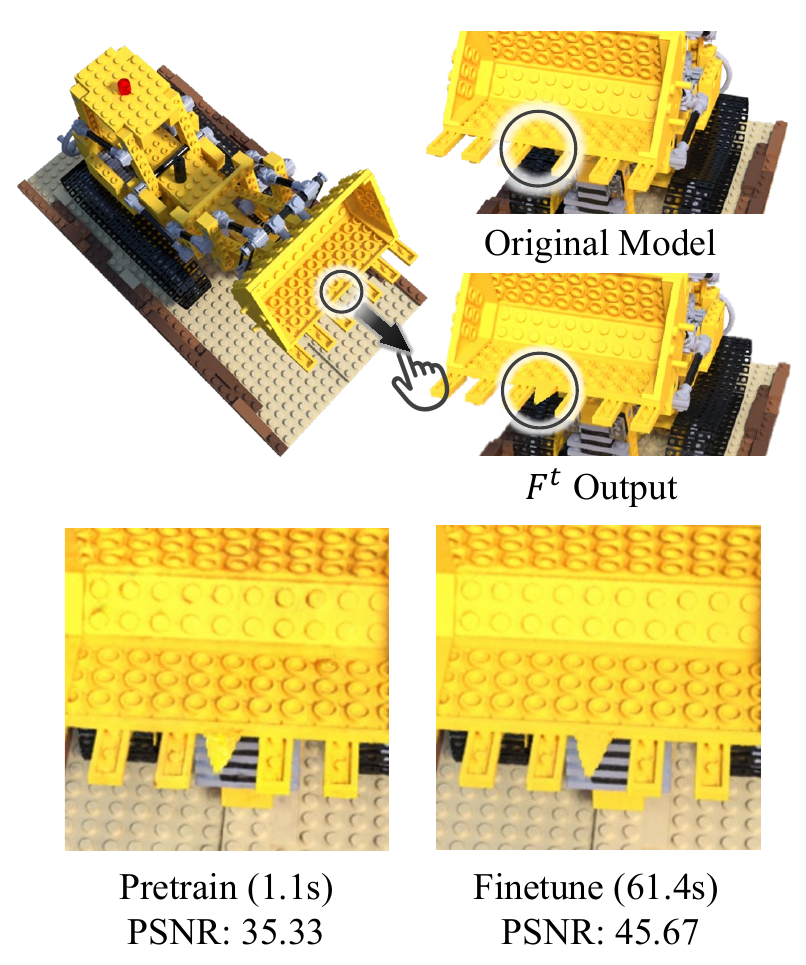}
    \caption{Example of anchor editing: fake tooth. }
    \label{fig-anchor}
\end{figure}

\begin{figure}
\centering
    \includegraphics[width=\linewidth, clip, trim=0 0px 0 0]{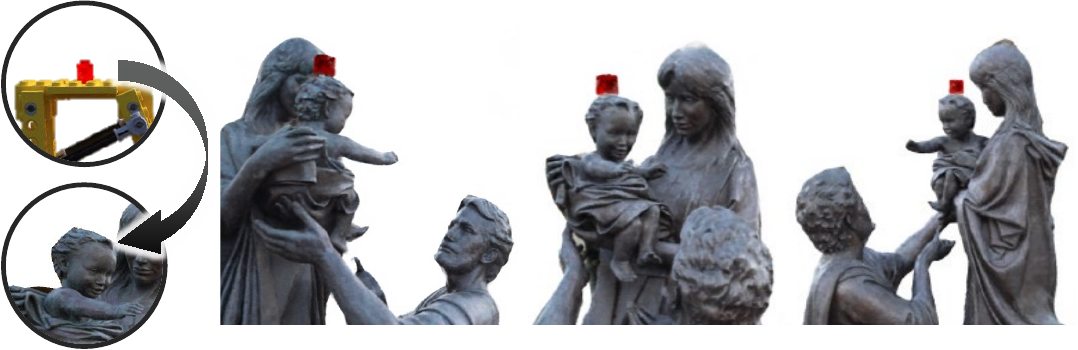}
    \label{fig-realresults-b}
    \vspace{-1.5em}
\caption{Example of object transfer editing: from the Lego scene (NeRF Blender) to the family scene (Tanks and Temples).}
\label{fig-bbox-baby}
\vspace{-1em}
\end{figure}

\subsection{Implementation Details}

\paragraph{Network.} In order to disentangle shape and color latent information within the hashgrids, we split the single hash table in the NeRF network architecture of Instant-NGP~\cite{mueller2022instant} into two: a density grid $\mathcal{G}^{\sigma}$ and a color grid $\mathcal{G}^c$, with the same settings as the original density grid in the open-source PyTorch implementation torch-ngp~\cite{torch-ngp}. We do this to make it possible to make fine-grained edits of one to one of the color or geometry properties without affecting the other. The rest of the network architecture remains the same, including a sigma MLP $f^\sigma$ and a color MLP $f^c$. For a spatial point $\mathbf{x}$ with view direction $\mathbf{d}$, the network predicts volume density $\sigma$ and color $c$ as follows:
\begin{align}
    \sigma, \mathbf{z} &= f^\sigma(\mathcal{G}^{\sigma}(\mathbf{x})) \\
    c &= f^c(\mathcal{G}^c(\mathbf{x}),\mathbf{z},\mathrm{SH}(\mathbf{d}))
\end{align}
where $\mathbf{z}$ is the intermediate geometry feature, and $\mathrm{SH}$ is the spherical harmonics directional encoder~\cite{mueller2022instant}. The same as Instant-NGP's settings, $f^\sigma$ has 2 layers with hidden channel 64, $f^c$ has 3 layers with hidden channel 64, and $\mathbf{z}$ is a 15-channel feature.

We compare our modified NeRF network with the vanilla architecture in the Lego scene of NeRF Blender Synthetic dataset\cite{mildenhall2020nerf}. We train our network and the vanilla network on the scene for 30,000 iterations. The result is as follows:
\begin{itemize}
    \item Ours: training time 441s, PSNR 35.08dB
    \item Vanilla: training time 408s, PSNR 34.44dB
\end{itemize}
We observe slightly slower runtime and higher quality for our modified architecture, indicating that this modification causes negligible changes.

\paragraph{Training.}
We select Instant-NGP~\cite{mueller2022instant} as the NeRF backbone of our editing framework.
Our implementations are based on the open-source PyTorch implementation torch-ngp~\cite{torch-ngp}. All experiments are run on a single NVIDIA RTX 3090 GPU. Note that we make a slight modification to the original network architecture. Please refer to the supplementary material for details.

During the pretraining stage, we set $\msymbol{weight_pretrain_color}=\msymbol{weight_pretrain_sigma}=1$ and the learning rate is fixed to $0.05$. During the finetuning stage, we set $\msymbol{weight_train_color} = \msymbol{weight_train_depth} = 1$ with an initial learning rate of 0.01. 
Starting from a pretrained NeRF model, we perform 50-100 epochs of local pretraining (for about 0.5-1 seconds) and about 50 epochs of global finetuning (for about 40-60 seconds). The number of epochs and time consumption can be adjusted according to the editing type and the complexity of the scene. Note that we test our performance in the absence of tiny-cuda-nn~\cite{tiny-cuda-nn} which achieves superior speed to our backbone, which indicates that our performance has room for further optimization.

\paragraph{Datasets.}
We evaluate our editing in the synthetic\Skip{lego, chair, and ship from} NeRF Blender Dataset~\cite{mildenhall2020nerf}, and the real-world captured \Skip{family and truck from}Tanks and Temples~\cite{Knapitsch2017} and \Skip{, and scan83 from} DTU~\cite{jensen2014large} datasets. We follow the official dataset split of the frames for the training and evaluation.

\begin{figure}
    \centering
    \includegraphics[width=\linewidth]{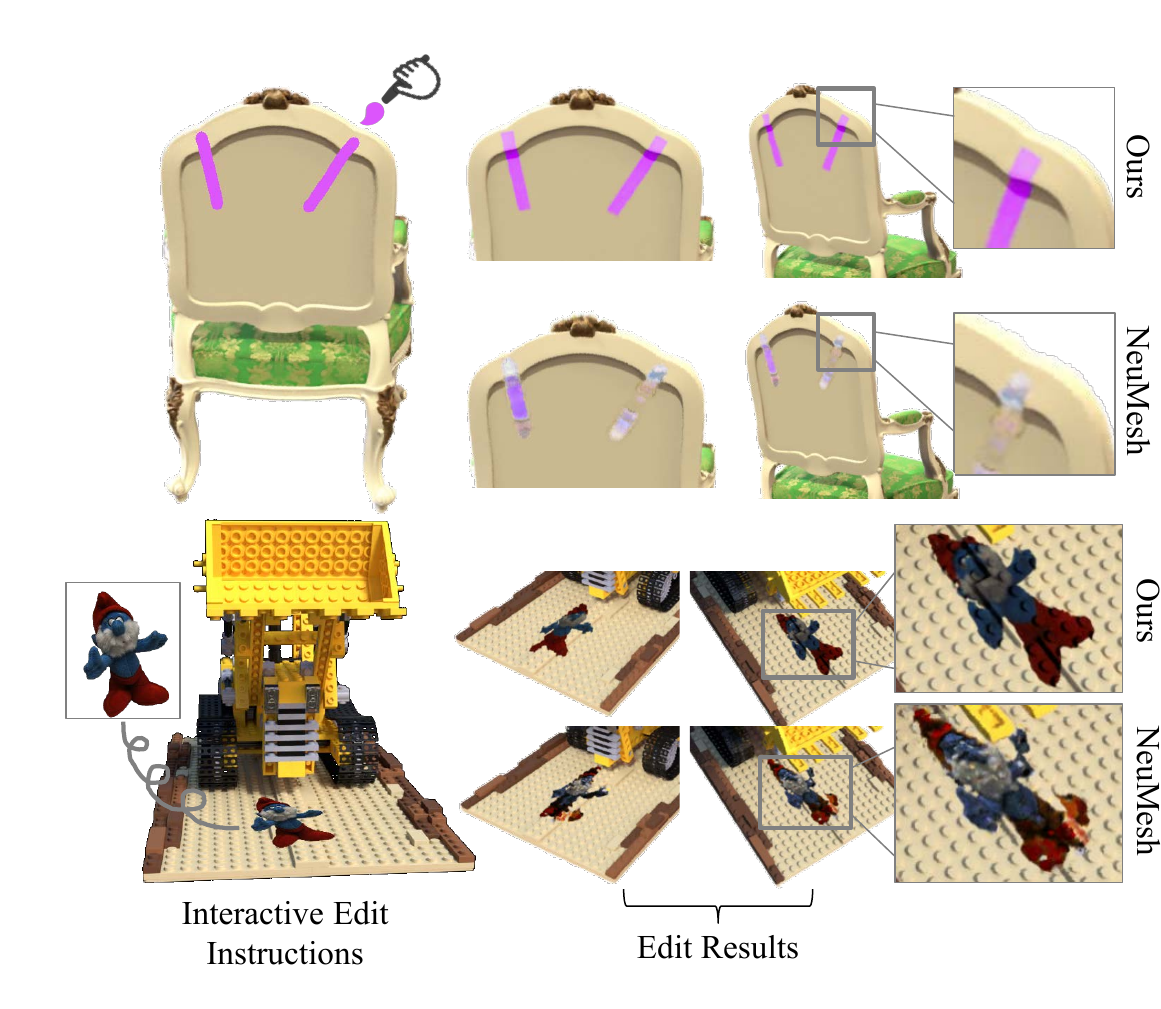}
    \vspace{-2em}
    \caption{Comparisons on texture/color painting between NeuMesh~\cite{neumesh} and our method. Note that NeuMesh requires hours of finetuning while ours needs only seconds.}
    \label{fig-neumesh}
\end{figure}

\begin{figure}
    \centering
    \includegraphics[width=\linewidth]{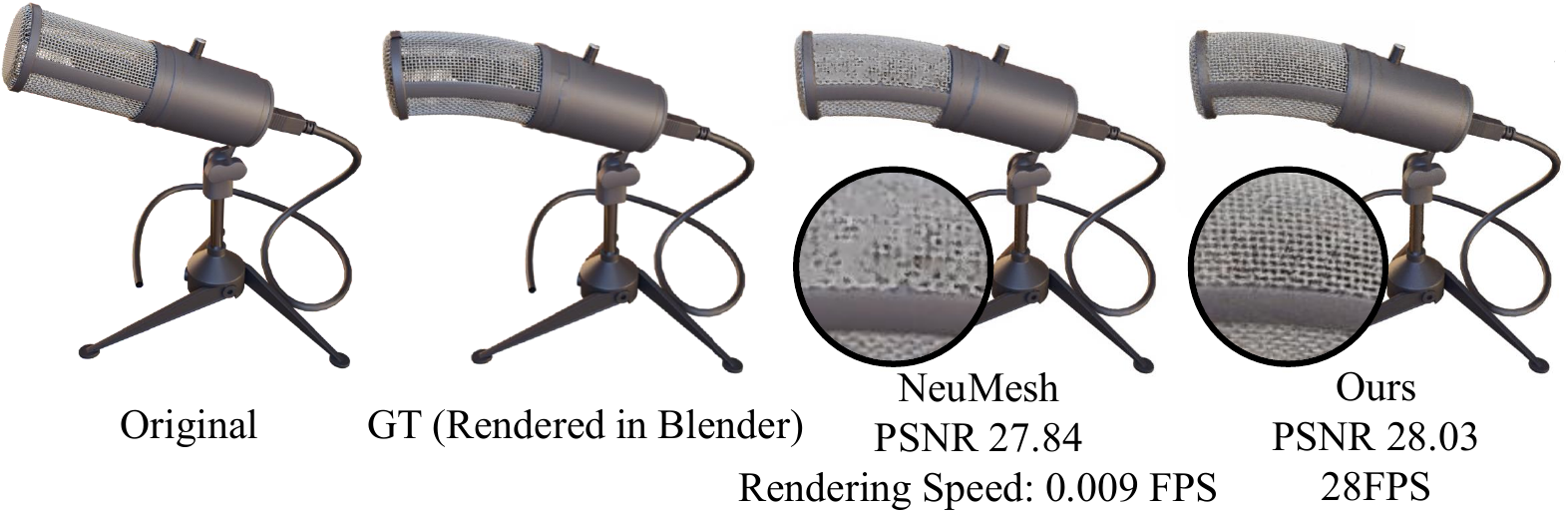}
    \caption{Comparison on qualitative and quantitative between NeuMesh \cite{neumesh} and our method.
    The PSNR is computed from the editing result and the rendering of the ground truth mesh with the same editing applied.}
    \label{fig-neumesh-mic}
\end{figure}

\begin{figure}
    \centering
    \setlength{\tabcolsep}{1pt}
    \vspace{-1em}
    \begin{tabular}{c|c|cc}
        \includegraphics[width=2.0cm,trim=100px 0 100px 40px,clip]{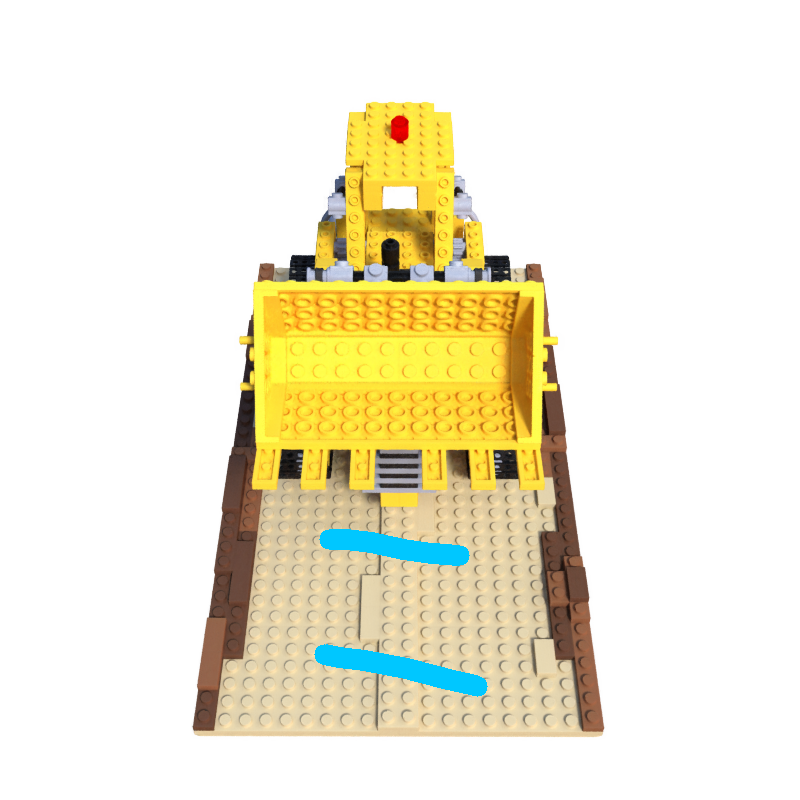} & \includegraphics[width=2.0cm,trim=80px 0 80px 40px,clip]{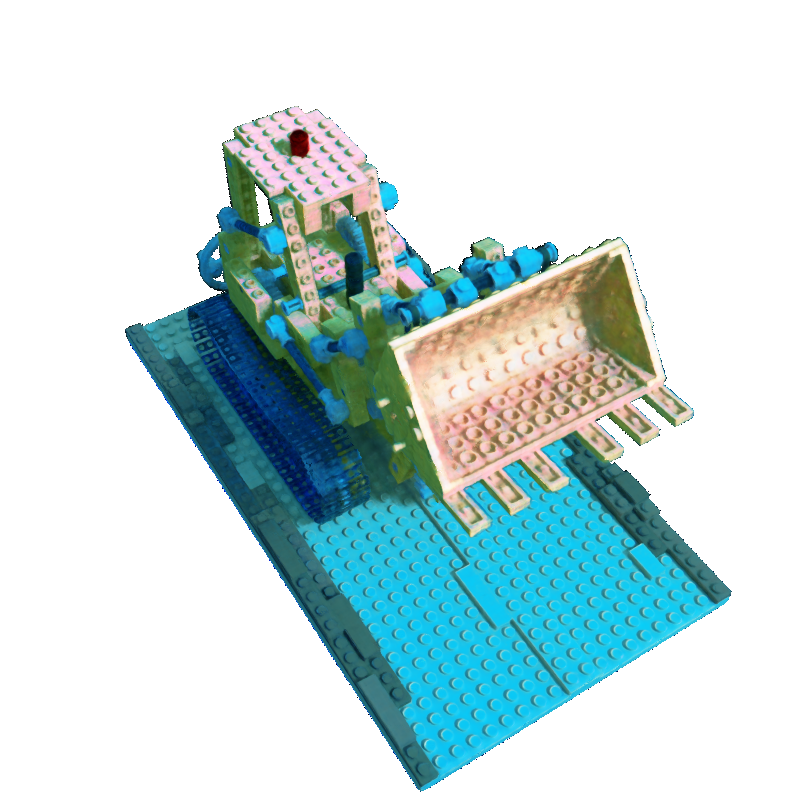} &
        \includegraphics[width=2.0cm,trim=80px 0 80px 40px,clip]{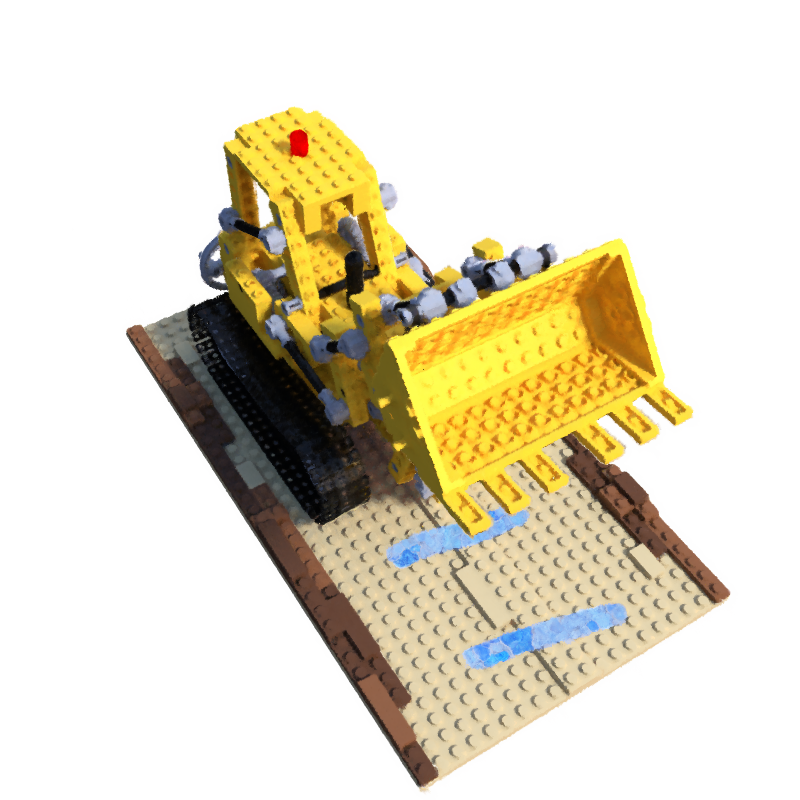}
        & \includegraphics[width=2.0cm,trim=80px 0 80px 40px,clip]{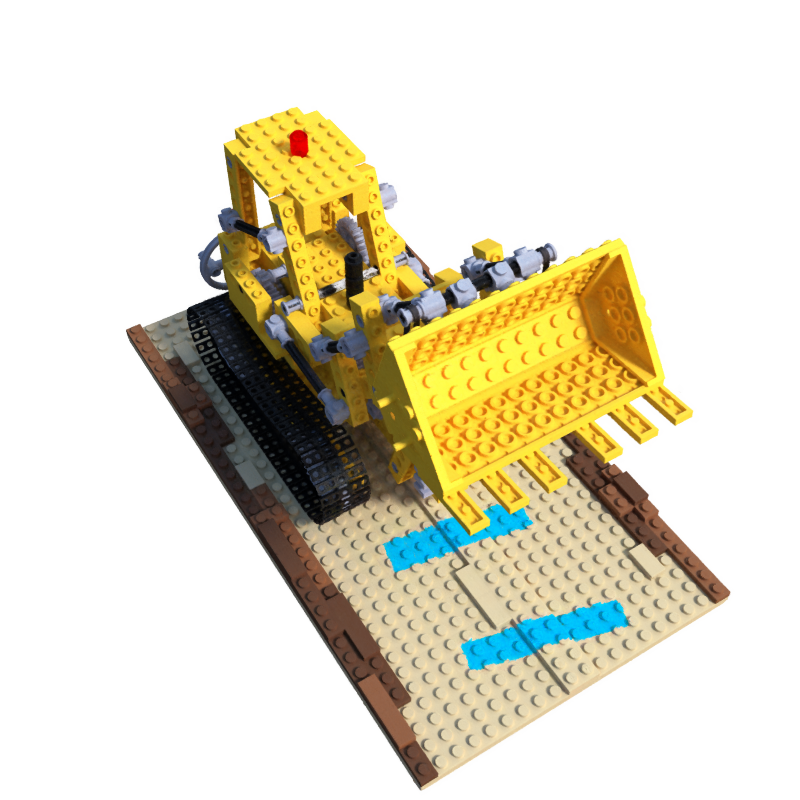} \\
        Instructions & Liu \etal~\cite{liu2021editing} & NeuMesh~\cite{neumesh} & Ours \\
    \end{tabular}
    \caption{Comparison of the pixel-wise editing ability between baselines~\cite{liu2021editing,neumesh} and ours. Note that \cite{liu2021editing} does not focus on the same task of pixel-wise editing as the other two. We are not to compete with their method.}
    \vspace{-1em}
    \label{fig-editnerf}
\end{figure}

\subsection{Experimental Results}
\label{sec-results}

\paragraph{Qualitative NeRF editing results.} 
We provide extensive experimental results in all kinds of editing categories we design, including bounding shape (\cref{fig-bbox,fig-bbox-elf}), brushing (\cref{fig-brush}), anchor (\cref{fig-anchor}), and color (\cref{fig-teaser}). Our method not only achieves a huge performance boost, supporting instant preview at the second level but also produces more visually realistic editing appearances, such as shading effects on the lifted side in \cref{fig-brush} and shadows on the bumped surface in \cref{fig-neumesh}. Besides, results produced by the student network can even outperform the teacher labels, \eg in \cref{fig-bbox-elf} the $F^t$ output contains floating artifacts due to view inconsistency. As analyzed in \cref{sec-train}, the distillation process manages to eliminate this. We also provide an example of object transfer (\cref{fig-bbox-baby}): the bulb in the Lego scene (of Blender dataset) is transferred to the child's head in the family scene of Tanks and Temples dataset.

\begin{figure}[ht]
    \centering
    \includegraphics[width=\linewidth]{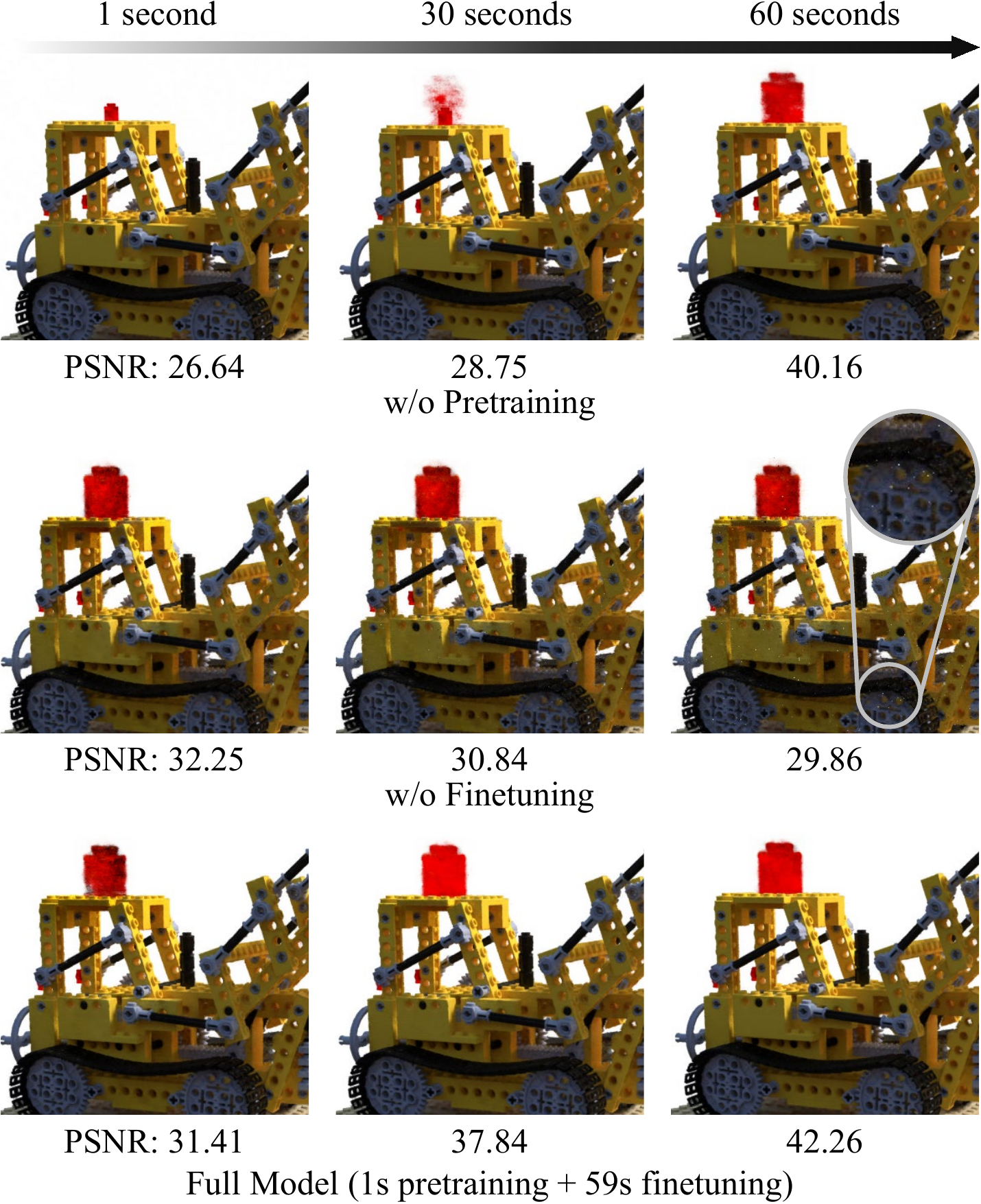}
    \caption{Ablation studies on two-stage training strategy. Zoom in for degradation details of ``w/o finetuning''.}
    \label{fig-ab_pre}
    \vspace{-1em}
\end{figure}

\begin{figure}[ht]
    \centering
    \includegraphics[width=\linewidth, trim=0 20px 0 0,clip]{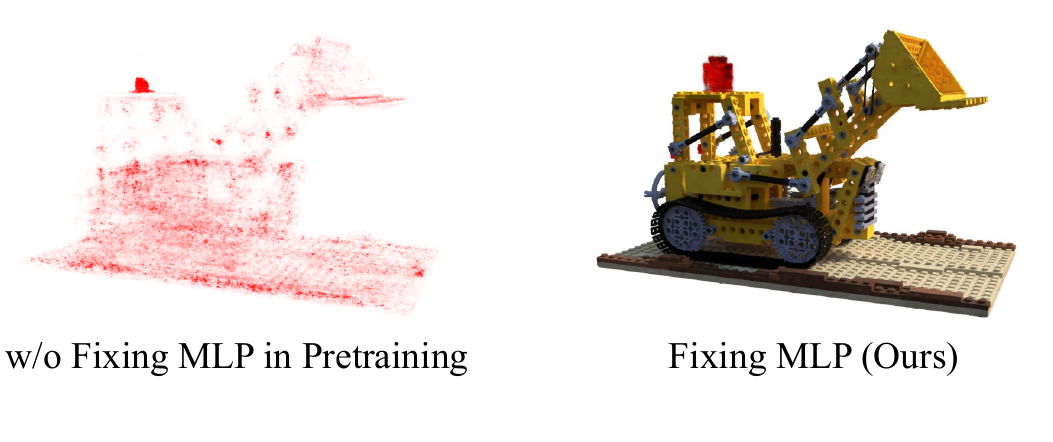}
    \caption{Ablation study on MLP fixing.}
    \label{fig-ab_fix}
    \vspace{-1.5em}
\end{figure}

\paragraph{Comparisons to baselines.} Existing works have strong restrictions on editing types, which focus on either geometry editing or appearance editing, while ours is capable of doing both simultaneously. Our brushing and anchor tools can create user-guided out-of-proxy geometry structures, which no existing methods support. We make comparisons on color and texture painting supported by NeuMesh~\cite{neumesh} and Liu \etal~\cite{liu2021editing}. 

\cref{fig-neumesh} illustrates two comparisons between our method and NeuMesh~\cite{neumesh} in scribbling and a texture painting task. Our method significantly outperforms NeuMesh, which contains noticeable color bias and artifacts in the results. In contrast, our method even succeeds in rendering the shadow effects caused by geometric bumps.

\cref{fig-neumesh-mic} illustrates the results of the same non-rigid blending applied to the Mic from NeRF Blender\cite{mildenhall2020nerf}. It clearly shows that being mesh-free, We have more details than NeuMesh\cite{neumesh}, unlimited by mesh resolution.

\cref{fig-editnerf} shows an overview of the pixel-wise editing ability of existing NeRF editing methods and ours. Liu \etal~\cite{liu2021editing}'s method does not focus on the pixel-wise editing task and only supports textureless simple objects in their paper. Their method causes an overall color deterioration within the edited object, which is highly unfavorable. This is because their latent code only models the global color feature of the scene instead of fine-grained local features. Our method supports fine-grained local edits due to our local-aware embedding grids.




\subsection{Ablation Studies}
\label{sec-ablation}
\paragraph{Effect of the two-stage training strategy.} To validate the effectiveness of our pretraining and finetuning strategy, we make comparisons between our full strategy (3\textsuperscript{rd} row), finetuning-only (1\textsuperscript{st} row) and pretraining-only (2\textsuperscript{nd} row) in \cref{fig-ab_pre}. Our pretraining can produce a coarse result in only 1 second, while photometric finetuning can hardly change the appearance in such a short period. The pretraining stage also enhances the subsequent finetuning, in 30 seconds our full strategy produces a more complete result. However, pretraining has a side effect of local overfitting and global degradation. Therefore, our two-stage strategy makes a good balance between both and produces optimal results.

\paragraph{MLP fixing in the pretraining stage.} In \cref{fig-ab_fix}, we validate our design of fixing all MLP parameters in the pretraining stage. The result confirms our analysis that MLP mainly contains global information so it leads to global degeneration when MLP decoders are not fixed.

\section{Conclusion}

We have introduced an interactive framework for pixel-level editing for neural radiance fields supporting instant preview. Specifically, we exploit the two-stage teacher-student training method to provide editing guidance and design a two-stage training strategy to achieve instant network convergence to obtain coarse results as a preview. Unlike previous works, our method does not require any explicit proxy (such as mesh), improving interactivity and user-friendliness. Our method also supports preserving shading effects on the edited surface. One limitation is that our method does not support complex view-dependent lighting effects such as specular reflections, and can not change the scene illumination, which can be improved by introducing intrinsic decomposition. Besides, our method does not handle the reconstruction failures (such as floating artifacts) of the original NeRF network.

\section*{ACKNOWLEDGEMENT}

This work was supported in part by the Fundamental Research Funds for the Central Universities; NSFC under Grants (62103372, 62088101, 62233013); the Key Research and Development Program of Zhejiang Province (2021C03037); Zhejiang Lab (121005-PI2101); Information Technology Center and State Key Lab of CAD\&CG, Zhejiang University. 

{\small
\bibliographystyle{ieee_fullname}
\bibliography{egbib}
}





\end{document}